\documentclass[11pt,a4paper]{article}
\PassOptionsToPackage{breaklinks}{hyperref}
\usepackage[hyperref]{mrp19}
\usepackage{times}
\usepackage{latexsym}
\usepackage{breakurl}

\usepackage{multirow}
\usepackage{breakurl}
\usepackage{colortbl}
\usepackage{xcolor}
\usepackage{graphicx}
\newenvironment{citemize}{\begin{list}{$\bullet$}{\topsep=\smallskipamount\itemsep=0pt\parsep=1pt\labelwidth=.5em}}{\end{list}}
\newenvironment{cenumerate}{\begin{list}{\labelenumi}{\usecounter{enumi}\topsep=0pt\itemsep=0pt\parsep=1pt\labelwidth=.7em}}{\end{list}}
\newenvironment{cenumeratew}{\begin{list}{\labelenumi}{\usecounter{enumi}\topsep=0pt\itemsep=0pt\parsep=1pt\labelwidth=1.2em}}{\end{list}}

\aclfinalcopy 

\newcommand\confname{CoNLL 2019}

\newcommand\taskname{Cross-Framework Meaning Representation Parsing}
\newcommand\taskshort{MRP 2019}

\title{{\'U}FAL MRPipe at MRP 2019:\\\large UDPipe Goes Semantic in the Meaning Representation Parsing Shared Task}

\author{Milan Straka \and Jana Strakov\'{a}\\
  Charles University \\
  Faculty of Mathematics and Physics \\
  Institute of Formal and Applied Linguistics \\
  {\tt \{straka,strakova\}@ufal.mff.cuni.cz} \\}

\date{}

\begin{document}
\maketitle

\begin{abstract}
  We present a system description of our contribution to the \confname~shared
  task, \taskname~(\taskshort). The proposed architecture is our first attempt
  towards a semantic parsing extension of the UDPipe 2.0, a lemmatization, POS
  tagging and dependency parsing pipeline.

For the MRP 2019, which features five formally and linguistically different
  approaches to meaning representation (DM, PSD, EDS, UCCA and AMR), we propose
  a uniform, language and framework agnostic graph-to-graph neural network
  architecture. Without any knowledge about the graph structure, and
  specifically without any linguistically or framework motivated features, our
  system implicitly models the meaning representation graphs.

  After fixing a human error (we used earlier incorrect version of provided
  test set analyses), our submission would score third in the competition
  evaluation. The source code of our system is available at \url{https://github.com/ufal/mrpipe-conll2019}.
\end{abstract}

\section{Introduction}

The goal of the \confname~shared task, \taskname~(\taskshort;
\citealp{Oep:Abe:Haj:19}) is to parse a raw, unprocessed sentence into its
corresponding graph-structured meaning representation.

The \taskshort~features five formally and linguistically different approaches to
meaning representation with varying degree of linguistic and structural
complexity:

\begin{citemize}
  \item \textbf{DM:} DELPH-IN MRS Bi-Lexical Dependencies \cite{DM2012},
  \item \textbf{PSD:} Prague Semantic Dependencies
    \cite{Hajic2012,Miy:Oep:Zem:14},
  \item \textbf{EDS:} Elementary Dependency Structures \cite{Oepen2006},
  \item \textbf{UCCA:} Universal Conceptual Cognitive Annotation \cite{UCCA2013},
  \item \textbf{AMR:} Abstract Meaning Representation \cite{AMR2013}.
\end{citemize}

In line with the shared task objective to advance uniform meaning
representation parsing across distinct semantic graph frameworks, we propose
a uniform, language and structure agnostic graph-to-graph neural network
architecture which models semantic representation from input sequences. The
system is an extension of the UDPipe 2.0, a~tagging, lemmatization and
syntactic tool \cite{UDPipe2018,UDPipeBERT2019}.

Our contributions are the following:

\begin{itemize}
  \item We propose a uniform semantic graph parsing architecture, which
    accommodates simple directed cyclic graphs, independently on the
    underlying semantic formalism.
  \item Our method does not use linguistic information such as structural
    constraints, dictionaries, predicate banks or lexical databases.
  \item We added a new extension to UDPipe 2.0, a lemmatization, POS tagging
    and dependency parsing tool. The semantic extension parses semantic graphs
    from the raw token input, making use of the POS and lemmas (but not syntax)
    from the existing UDPipe 2.0.
  \item As an improvement over UDPipe 2.0, we use the ``frozen'' contextualized
    embeddings on the input (BERT; \citealp{BERT2019}) in the same way as
    \citet{UDPipeBERT2019}.
\end{itemize}

After fixing a human error (we used earlier incorrect version of provided test
set analyses), our submission would score third in the competition evaluation.

\section{Related Work}


Numerous parsers have been proposed for parsing semantic formalisms, including
the systems participating in recent semantic parsing shared tasks SemEval 2016
and SemEval 2017 \cite{SemEval2016,SemEval2017} featuring AMR; and SemEval 2019
\cite{SemEval2019} featuring UCCA. However, proposals of general, formalism
independent semantic parsers are scarce in the literature.

\citet{Hershovich2018} propose a general transition-based parser for directed,
acyclic graphs, able to parse multiple conceptually and formally different
schemes. TUPA is a transition-based top-down shift-reduce parser, while ours,
although also based on transitions/operations, models the graph as a sequence
of layered, iterative graph-like operations, rather (but not necessarily) in
a bottom-up fashion. Consequently, our architecture allows parsing cyclic
graphs and is not restricted to single-rooted graphs. Also, we do not enforce
any task-specific constraints, such as restriction on number of parents in UCCA
or number of children given by PropBank in AMR and we completely rely on the
neural network to implicitly infer such framework-specific features.

\section{Methods}

\subsection{Uniform Graph Model}

The five shared task semantic formalisms differ notably in specific formal and
linguistic assumptions, but from a higher-level view, they universally
represent the full-sentence semantic analyses with directed, possibly cyclic
graphs. Universally, the semantic units are represented with graph nodes and
the semantic relationships with graph edges.

To accommodate these semantic structures, we model them as directed simple
graphs $G = (V,E)$, where $V$ is a set of nodes and $E \subseteq \{(x,y) \,|\, (x,y)
\in V^2, x \neq y\}$ is a set of directed edges.\footnote{Specifically, our
graphs are directed and allow cycles. Furthermore, they are \textit{simple}
graphs, not \textit{multigraphs}.
}

One of the most fundamental differences between the five featured
\taskshort~frameworks lies apparently in the relationship between the graph
structure (graph nodes) and the input surface word forms (tokens). In the
\taskshort, this relationship is called \textit{anchoring} and its degree
varies from a tight connection between graph nodes being directly corresponding
to surface tokens in \textit{Flavor 0} frameworks (DM and PSD) through more
relaxed relationship \textit{Flavor 1} (EDS and UCCA) in which arbitrary parts
of the sentence can be represented in the semantic graph, to a completely
\textit{unanchored} semantic graph of \textit{Flavor 2} in the AMR framework.

To alleviate the need for a framework-specific handling of the anchoring, we
broaden our understanding of the semantic graph: We consider the tokens as
nodes and the anchors (connections from the graph nodes to tokens) as
regular edges, thus the anchors are naturally learned jointly with the graph
without an explicit knowledge of the underlying semantic formalism.

In order to represent anchors as regular edges in the graph, the input
tokenization needs to be consistent with the annotated anchors: each anchor
must match one or multiple input tokens. In order to achieve the exact
anchor-token(s) match, we created a simple tokenizer. The tokenizer is uniform
for all frameworks with a slight change to capture UCCA's fine-grained
anchoring; see Figure~\ref{fig:tokenizer} for the
pseudocode.\footnote{Instead of generating tokens consistent with the anchors,
the anchoring edges could be allowed to refer only to a part of a token (for
example by having two attributes \textit{first anchored token character} and
\textit{last anchored token character}), which is an approach we plan to adopt
in the future.}

Furthermore, to represent anchors as edges, the anchors have to be annotated in
the data, which is not the case for AMR. We therefore utilize externally
generated anchoring from the JAMR tool \cite{JAMR2016}.\footnote{We plan to
model the anchors jointly using an attention mechanism \cite{Zhang2019}.}

\begin{figure*}
  \small
  \begin{cenumeratew}
    \item[1.] Any single non-space character
    \item[\textcolor{gray}{2a.}] {\color{gray}UCCA: \verb|\w+[$]?|}
    \item[2b.] other: \verb+\w(\w-[^-\s]|&|/|'S\w|'[A-RT-Z]|[.](?=.*\w)\w|\d)*[$]?+;\\
      \verb|\d+-\d+|; \verb|\d+,\d+|; \verb|\d+,\d+,\d+|
    \item[3.] \verb|--+|; \verb|`+|; \verb|'+|; \verb|[.]+|; \verb|!+|
    \item[4.] \verb|n't|; \verb+'s+;
      \verb+'d+ ; \verb+'m+ ; \verb+'re+ ; \verb+'ve+ ; \verb+'ll+
    \item[5.] Split the following word into two tokens:
      \verb+would|n't+; \verb+could|n't+; \verb+ca|n't+; \verb+is|n't+;
      \verb+are|n't+; \verb+ai|n't+; \verb+was|n't+; \verb+were|n't+;
      \verb+do|n't+; \verb+does|n't+; \verb+did|n't+; \verb+should|n't+;
      \verb+have|n't+; \verb+has|n't+; \verb+had|n't+; \verb+wo|n't+;
      \verb+might|n't+; \verb+need|n't+; \verb+can|not+; \verb+wan|na+;
      \verb+got|ta+
  \end{cenumeratew}
  \caption{Tokenizer pseudocode as a sequence of regular expressions.
  Expressions with higher number override previous ones.}
  \label{fig:tokenizer}
\end{figure*}

\subsection{Graph-to-graph Parser}

\begin{figure*}
  \includegraphics[width=\hsize]{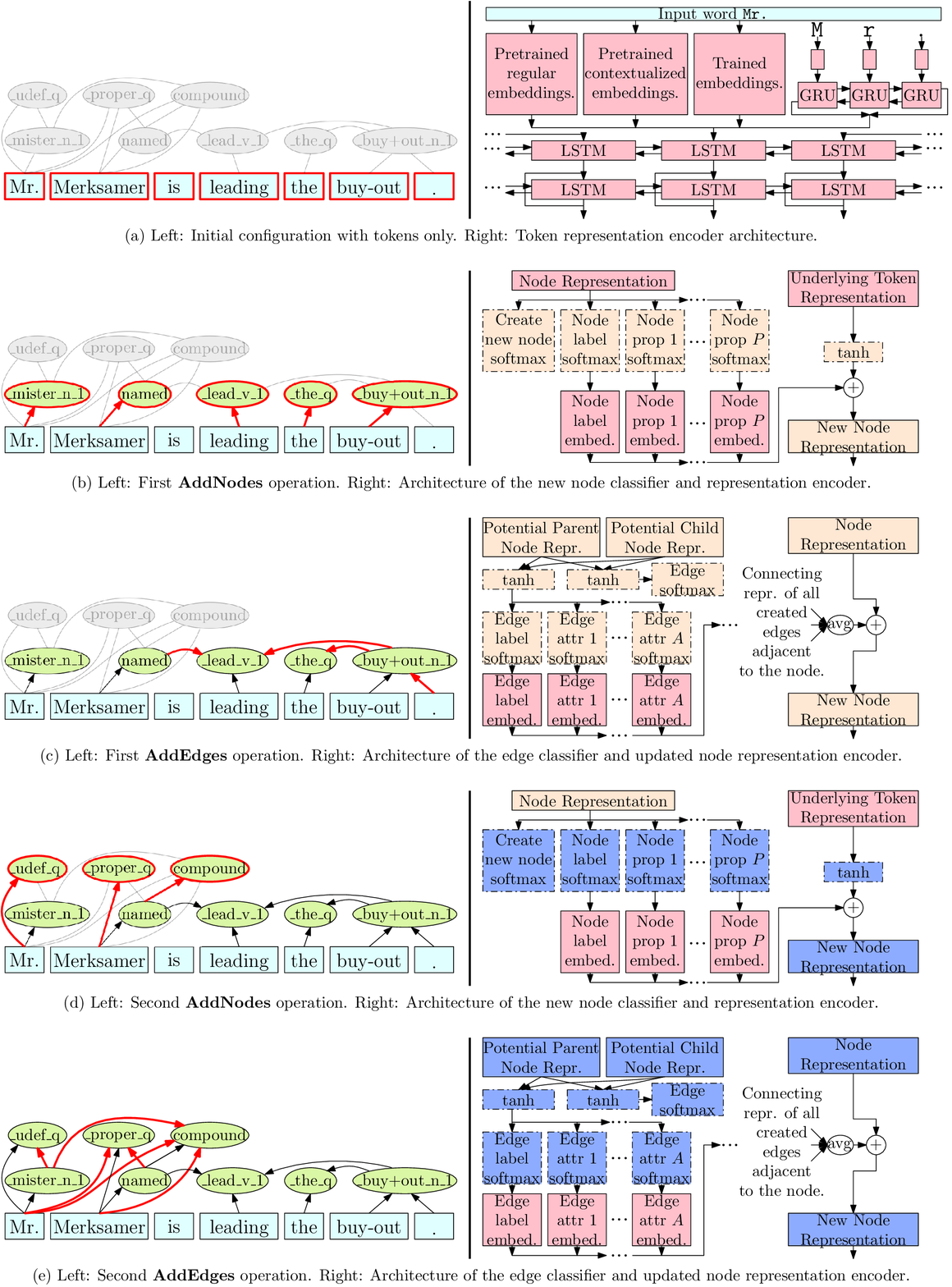}
  \caption{Our graph-to-graph architecture schematic overview and an example of
  semantic graph build-up for the sentence \textit{``Mr. Merksamer is leading the
  buy-out.''} from the EDS framework \cite{Oepen2006}. Note that the weights
  for all classification layers and for all displayed fully connected layers
  (displayed with dashed border) are different for every iteration of
  \textbf{AddNodes}/\textbf{AddEdges} operations.}
  \label{fig:schema}
\end{figure*}

We propose a general graph-to-graph parser which models the graph meaning
representation as a sequence of layered group transformations from input from
input sequence to meaning graphs. A~schematic overview of our architecture is
presented in Figure~\ref{fig:schema}.

Having reduced the task to a graph-to-graph transformation modeling, we
iteratively build the graph from its initial state (a set of isolated nodes --
tokens) by alternating between two layer-wise transformations:

\begin{cenumerate}
  \item \textbf{AddNodes}: The first operation creates new nodes and connects them
    to already existing nodes. Specifically, for each already existing node we
    decide whether to a) create a new node and connect it as a parent,
    b) create a new node and connect it as a child, c) do nothing. When a new
    node is created, its label and all its properties are generated too.

    Intuitively, anchors are modeled in the first step from the initial set of
    individual nodes (tokens) and in the next steps, higher-layer nodes are
    modeled. As a special case, \textbf{AddNodes} is relatively simple for
    the Flavor 0 frameworks (DM and PSD): zero or one node is created for
    every token in the first and only \textbf{AddNodes} iteration.
    This is illustrated in Table~\ref{tbl:coverage}, which shows node coverage
    after performing a fixed number of \textbf{AddNodes} iterations, reaching
    $100\%$ after one \textbf{AddNodes} iteration in DM and PSD.

  \item \textbf{AddEdges}: The second operation creates edges between the new
    nodes and any other existing nodes (both old and new) using a~classifier for
    each pair of nodes. Any number of edges can be connected to a newly created
    node.
\end{cenumerate}

At the end of each iteration, the created nodes and edges are frozen and
the computation moves to its next iteration. We describe the crucial part of
the graph modeling, \textbf{token}, \textbf{node} and \textbf{edge
representation}, in Section~\ref{section:encoder}.

An example of a graph step by step build-up is shown in
Figure~\ref{fig:schema}.

In contrast to purely sequential series of single transitions, such as adding
a new edge in one step, adding new nodes and edges in a layer-wise fashion
improves runtime performance and might avoid error accumulation by performing many
independent decisions. On the other hand, we assume that creating nodes from
a single existing one might be problematic, especially if the graph has constituency
structure.

\subsubsection{Creating \textbf{AddNodes} Operations}

\begin{table*}
  \centering
  \small
  \setlength{\tabcolsep}{5pt}
  \begin{tabular}{l|l*{10}{|c}}
    \multicolumn{2}{c|}{\multirow{2}{*}{Framework}} & \multicolumn{10}{c}{Iterations} \\\cline{3-12}
    \multicolumn{2}{c|}{} & 1 & 2 & 3 & 4 & 5 & 6 & 7 & 8 & 9 & 10 \\\hline
    \multirow{2}{*}{DM}   & Nodes & \llap{1}00.00\% \\\cline{2-12}
                          & Graphs & \llap{1}00.00\% \\\hline
    \multirow{2}{*}{PSD}  & Nodes & \llap{1}00.00\% \\\cline{2-12}
                          & Graphs & \llap{1}00.00\% \\\hline
    \multirow{2}{*}{EDS}  & Nodes &  69.18\% & 97.18\% & 99.31\% & 99.64\% & 99.90\% & 99.95\% & 99.97\% & 99.99\% & \llap{1}00.00\% & \llap{1}00.00\%\\\cline{2-12}
                          & Graphs & ~~2.15\% & 59.31\% & 91.68\% & 93.09\% & 98.84\% & 99.46\% & 99.55\% & 99.87\% & 99.99\% & 99.99\% \\\hline
    \multirow{2}{*}{UCCA} & Nodes & 69.63\% & 97.57\% & 99.87\% & 99.97\% & \llap{1}00.00\% & \llap{1}00.00\% \\\cline{2-12}
                          & Graphs & ~~0.00\% & 43.72\% & 97.29\% & 99.19\% & 99.92\% & \llap{1}00.00\% \\\hline
    \multirow{2}{*}{AMR}  & Nodes & 78.23\% & 96.15\% & 99.01\% & 99.69\% & 99.88\% & 99.94\% & 99.96\% & 99.97\% & 99.98\% & 99.99\% \\\cline{2-12}
                          & Graphs & 19.73\% & 74.58\% & 93.48\% & 98.18\% & 99.49\% & 99.86\% & 99.94\% & 99.96\% & 99.97\% & 99.98\% \\\hline
  \end{tabular}
  \caption{Coverage of training graphs after a fixed number of the layer-wise iterations. Rows labeled ``Nodes'' show percentage of covered nodes. Rows labeled ``Graphs'' show percentage of complete graphs.}
  \label{tbl:coverage}
\end{table*}

For training, a sequence of the \textbf{AddNodes} operations must be created.
For this purpose, we define an ordering of the graph nodes which guides the
graph traversal. The initial order of the isolated graph nodes set (tokens) is
left to right, the first token being the first to be visited. The other graph
nodes' ordering is then induced by the order of creation.

\looseness1
Given a training graph, we then generate a sequence of \textbf{AddNodes}
operations. In every iteration, we traverse all existing nodes in the graph in
the above defined order and for each node, we consider all its not-yet-created
neighbors, from which we choose the one which is ``in the lowest layer''. This
is motivated by our intention to build the graph in a bottom-up fashion.
Specifically, we choose such a~node which has the smallest number of token
descendants (based on the assumption that nodes in the lower levels tend to
govern less descendants than the nodes in the higher levels), and if there are
several such nodes, the one where the token descendant indices are smallest in
the ordering. Finally, we favour creating parents to creating children, and
if a node can be created as a parent, we never create it as a~child.

As a special case, the first iteration always traverses the set of isolated
nodes (tokens) and connects their immediate parents with the anchor-defined
edges. For DM and PSD frameworks, this is the first and only iteration of the
\textbf{AddNodes} operations.

The number of required iterations to generate all nodes and construct complete
graphs is presented in Table~\ref{tbl:coverage}. Performing three iterations is
enough to cover more than 99\% of nodes in all frameworks, but EDS and AMR
frameworks sometimes require more than 10 iterations to generate a full graph.

During inference, we currently perform a fixed number of iterations of
\textbf{AddNodes} and \textbf{AddEdges} operations; we use one iteration for DM
and PSD, two iterations for UCCA and AMR, and three iterations for EDS.
Alternatively, we could allow a dynamic number of iterations, stopping when
\textbf{AddNodes} generates no new nodes.


\subsection{Node Labels and Properties Encoding}

Besides the graph structure, node labels and properties must also be modeled.
For some node labels or properties, it might be beneficial to generate them
relatively to a token. For example, when creating a lemma \textit{look} from
a token \textit{looked}, it might be easier to generate it as a rule
\textit{remove the last two token characters} instead of generating
\textit{look} directly. Such approach was taken by UDPipe lemmatizer
\cite{UDPipe-SIGMORPHON2019}, which produced the best results in lemmatization
in Task 2 of the SIGMORPHON 2019 Shared Task.

We adopt this approach, and generate all node labels and properties using
a simple classification into a collection of rules. Each rule can either
generate an independent value (which we call \textit{absolute encoding})
or it describes how a value should be created from a token (which we call
\textit{relative encoding}). For detailed description of the relative encoding
rules, please refer to \citet{UDPipe-SIGMORPHON2019}. In short, the lemmas in
UDPipe are generated by classifying into a set of character edit
scripts performed on the prefix and suffix. First,
a common root is found between the input and the output (word form and lemma).
If there is no common character, the lemma is
considered irregular and an \textit{absolute encoding} is used. Otherwise,
the shortest edit script is computed for the prefix and suffix.

\begin{table}[t]
  \centering
  \small
  \def\x{\bfseries}
  \def\!{\kern-.4em }
  \begin{tabular}{l|l|r|r}
    \multirow{2}{*}{\!Framework\!} & \multirow{2}{*}{\!Property\!} & \multicolumn{1}{c|}{Absolutely} & \multicolumn{1}{c}{Relatively} \\
                                       &                           & \!encoded values\! & \!encoded values\! \\\hline\hline
    \multirow{3}{*}{DM}   & label & 26\,907 &\x 1\,086 \\\cline{2-4}
                          & pos &\x 38 & 356 \\\cline{2-4}
                          & frame &\x 468 & 2613 \\\hline
    \hline
    \multirow{3}{*}{PSD}  & label & 32\,284 &\x 774 \\\cline{2-4}
                          & pos &\x 42 & 314 \\\cline{2-4}
                          & frame &\x 5\,294 & 8\,868 \\\hline
    \hline
    \multirow{2}{*}{EDS}  & label & 15\,905 &\x 4\,339 \\\cline{2-4}
                          & carg & 13\,667 &\x 427 \\\hline
    \hline
    \multirow{1}{*}{UCCA} & --- & --- & --- \\\hline
    \hline
    \multirow{17}{*}{AMR}  & label & 14\,554 &\x 6\,278 \\\cline{2-4}
                          & op1 & 7\,377 &\x 1\,402 \\\cline{2-4}
                          & op2 & 3\,673 &\x 545 \\\cline{2-4}
                          & op3 & 1\,149 &\x 242 \\\cline{2-4}
                          & op4 & 482 &\x 113 \\\cline{2-4}
                          & op5 & 245 &\x 56 \\\cline{2-4}
                          & ARG1 & 48 &\x 30 \\\cline{2-4}
                          & ARG2 & 127 &\x 68 \\\cline{2-4}
                          & ARG3 & 22 &\x 20 \\\cline{2-4}
                          & quant & 885 &\x 603 \\\cline{2-4}
                          & value & 861 &\x 590 \\\cline{2-4}
                          & time &\x 110 & 111 \\\cline{2-4}
                          & year & 153 &\x 58 \\\cline{2-4}
                          & li & 56 &\x 40 \\\cline{2-4}
                          & mod & 79 &\x 33 \\\cline{2-4}
                          & day &\x 31 & 57 \\\cline{2-4}
                          & month &\x 14 & 17 \\\cline{2-4}
                          & \ldots & \ldots & \ldots \\
  \end{tabular}
  \caption{Cardinality of absolute and relative encoded node properties in all
  frameworks. The chosen encoding is displayed in \textbf{bold}.}
  \label{tbl:encoding}
\end{table}

In our setting, however, we need to extend the UDPipe approach in two
directions. First, some properties like \textit{pos} should never be relatively
encoded. Therefore, during data loading, we consider both allowing and
disallowing relative encoding, and choose the approach yielding the smaller
number of classes. As Table~\ref{tbl:encoding} indicates, even such a simple
heuristic seems satisfactory.

Second, compared to lemmatization, where the lemma and the original form are
single words, in our setting both the property and the anchored tokens can be
a sequence of words (e.g., ``Pierre Vinken''). We overcome this issue by
encoding each word of a property independently, and for every property word, we
choose a subsequence of anchoring tokens which yields the shortest relative
encoding.

%

\subsection{Graph Representation}
\label{section:encoder}

\textbf{Token Encoder.} The input representation is a sequence of tokens
encoded as a concatenation of word and character-level word vectors:

\begin{citemize}
  \item trainable word embeddings (WE),
  \item character-level word embeddings (CLE): bidirectional GRUs in
    line with \citet{Ling2015}. We represent every Unicode character with
    a vector of dimension $256$, and concatenate GRU output for forward and
    reversed word characters. The character-level word embeddings are trained
    together with the network.
  \item pre-trained FastText word embeddings of dimension $300$
    \cite{FastText2018},\footnote{\url{https://fasttext.cc/docs/en/english-vectors.html}}
  \item pre-trained (``frozen'') contextual BERT embeddings of
    dimension $768$ \cite{BERT2019}.\footnote{We use the
    {\footnotesize\ttfamily Base English Uncased} model from
    \url{https://github.com/google-research/bert}.} We average the last four
    layers of the BERT model and we produce a word embedding for a token as an
    average of the corresponding BERT subword embeddings.

    Contextualized embeddings have recently been shown to improve performance of
    many NLP tasks, see for example \citet{UDPipeBERT2019} in the context of
    UDPipe and POS tagging, lemmatization and dependency parsing. Therefore,
    we expected that utilization of BERT embeddings would improve results
    considerably, which was the case, as demonstrated in
    Section~\ref{sec:ablation}.
\end{citemize}

Furthermore, the input tokens could be processed by a POS tagger, lemmatizer,
dependency parser or a named entity recognizer. If such analyses are available,
they can be used as additional embeddings of input tokens. Specifically,
we utilize the POS tags and lemmas provided in the shared task. We did
not experiment with dependency parses, which we plan to do in the
future. Furthermore, we tried utilizing the Illinois Named Entity Tagger
\cite{RatinovRo09}, but it did not improve our results.

All available embeddings for a token are concatenated and processed with two
bidirectional LSTM layers with residual connections.

\noindent\textbf{Node Encoder.} A node is represented by a concatenation of
these features:

\begin{citemize}
  \item the (transitively) attaching token representation (every node has
    exactly one token which generated it using the \textbf{AddNodes}
    operations), transformed by a dense layer followed by $\tanh$ nonlinearity;
    every \textbf{AddNodes} iteration has its own dense layer weights,
  \item the node \textit{label} and \textit{properties} embeddings,
  \item an average of edge representations of all connected edges.
\end{citemize}

A natural extension would be to represent all node's descendants instead of the
one token generating this node through a sequence of \textbf{AddNodes}, because
the current implementation seems to generate suboptimal representations in
later iterations. We leave a proper way of propagating all information through
the graph as our future work.

\noindent\textbf{Edge Representation.} An edge is represented by a sum
of its \textit{label} and \textit{attributes} embeddings.

\subsection{Decoders}

In the \textbf{AddNodes} operation, we employ the following classification
decoders, each utilizing the node representation and consisting of a fully
connected layer followed by a softmax activation:
\begin{citemize}
  \item decide among three possibilities, whether to a) add a node as a parent,
    b) add a node as a child, or c) do nothing;
  \item generate node label;
  \item for each property, generate its value (or a special class \verb|NONE|).
\end{citemize}

During training, we sum the losses of the decoders, apart from the situation
when no new node is created, in which case we ignore the label and properties
losses.

In the \textbf{AddEdges} operation, we consider all edges to and from the newly
created nodes. Utilizing all suitable pairs of nodes, we decide for each pair
separately whether to add an edge or not.

Although biaffine attention seems to be the preferred architecture for
dependency parsing recently \cite{udst2018:overview}, in our experiments
it performed poorly when we used it for deciding whether to add an edge between
any pair of nodes individually.
Our hypothesis is that the range of the biaffine attention output is changing
rapidly. That is not an issue when the outputs ``compete'' with each other in
a softmax layer, but is problematic when we compare each with a fixed
threshold.

Consequently, we utilized a Bahdanau-like attention \cite{bahdanau14} instead.
Specifically, we pass potential parent and child nodes' representations through
a pair of fully connected layers with the same output dimensionality, sum the results,
apply a $\tanh$ nonlinearity, and attach a binary classifier (a fully connected
layer with two outputs and a softmax activation) indicating whether the edge
should be added.\footnote{We always add an edge generated in the \textbf{AddNodes}
operation independently on the prediction for that edge in the \textbf{AddEdges}
operation.}

In order to predict edge label and attributes, we repeat the same attention
process (pass potential parent and child nodes' representation through
a different pair of fully connected layers, sum and $\tanh$), and attach
classifiers for edge labels and as many edge attributes as present in the data.

Lastly, in order to predict \textit{top} nodes, we employ a sigmoid binary
classifier processing the final node representations.

Finally, every iteration of \textbf{AddNodes} and \textbf{AddEdges} operations
has invididual set of weights for all layers described in this section.

\subsection{Training}

We implemented the described architecture using TensorFlow 2.0 beta
\cite{tensorflow_eager}. The
eager evaluation allowed us to construct inputs to \textbf{AddNodes} and
\textbf{AddEdges} for every batch specifically, so we could easily handle
dynamic graphs.

We trained the network using a lazy variant of Adam optimizer
\cite{Adam}\footnote{\texttt{tf.contrib.opt.lazyadamoptimizer}} with
$\beta_2=0.98$, for 10 epochs with a learning rate of $10^{-3}$ and for 5 additional
epochs with a learning rate $10^{-4}$ (the difference being UCCA which used 15 and
10 epochs, respectively, because of considerably smaller training data). We
utilized a batch size of 64 graphs.\footnote{Because we trained on a 8GB GPU,
we actually needed to process two batches of size 32 and only then perform
parameter update using summed gradients.} The training time on a single GPU was
1-4 hours for DM, PSD, EDS and UCCA, and 10 hours for AMR.

For replicability, we also describe the used hyperparameters in detail. The only
differences among the frameworks were:
\begin{citemize}
  \item slightly different tokenizer for UCCA (Fig~\ref{fig:tokenizer}),
  \item larger number of training epochs for UCCA,
  \item number of layer-wise iterations: 1, 1, 3, 2, 2 for DM, PSD, EDS, UCCA
    and AMR, respectively.
\end{citemize}
In the encoder, we utilized trainable embeddings of dimension 512, and
trainable character-level embeddings using character embeddings of size 256
and a single layer of bidirectional GRUs with 256 units. We processed token
embeddings using two layers of bidirectional LSTMs with residual connections
and a dimension of 768. The node representations also had dimensionality
768, as did node label and properties embeddings. We employed dropout with rate
0.3 before and after every LSTM layer and on all node representations, and
utilized also word dropout (zeroing the whole WE for a given word) with a rate
of 0.2. In the \textbf{AddEdges} operation, all attention layers have
a~dimensionality of 1024.

\subsection{Data Preprocessing}

We created two train/dev splits from the training data provided by the
organizers: Firstly, a $90\%/10\%$ train/dev split was used to train the model
and tune the hyperparameters of the competition entry. For the ablation
experiments in the post-competition phase, we later tried a $99\%/1\%$
train/dev split, which improved the results only marginally, as shown in
Section~\ref{sec:ablation}.

We further used the provided morphological annotations and the JAMR anchoring
for the AMR framework \cite{JAMR2016}.

\section{Results}
\label{section:results}

\begin{table*}
  \centering
  \small\setlength{\tabcolsep}{4.5pt}
  \begin{tabular}{l|r|r|r|r|r|r|r}
System & \multicolumn{1}{c|}{Tops} & \multicolumn{1}{c|}{Labels} & \multicolumn{1}{c|}{Properties} & \multicolumn{1}{c|}{Anchors} & \multicolumn{1}{c|}{Edges} & \multicolumn{1}{c|}{Attributes} & \multicolumn{1}{c}{All}\\\hline\hline
Original ST submission & 75.12\% ~~6 & 63.99\% ~~7 & 56.53\% ~~6 & 69.53\% ~~6 & 62.17\% ~~7 & 7.85\% ~~4 & 74.74\% ~~6\\\hline
Bugfix ST submission & 81.47\% ~~6 & \bfseries73.06\% ~~1 & \bfseries69.95\% ~~1 & 77.23\% ~~3 & 73.89\% ~~5 & 7.87\% ~~4 & 83.96\% ~~3\\\hline
99\% training data & 80.59\% ~~6 & \bfseries73.06\% ~~1 & \bfseries70.18\% ~~1 & 77.35\% ~~3 & 74.27\% ~~5 & 7.96\% ~~4 & 84.14\% ~~3\\\hline
No BERT embeddings & 70.50\% ~~8 & 70.71\% ~~4 & 67.01\% ~~4 & 76.02\% ~~4 & 65.02\% ~~6 & 5.30\% ~~6 & 78.99\% ~~5\\\hline
Ensemble & 81.13\% ~~6 & \bfseries\cellcolor[gray]{0.8}73.39\% ~~1 & \bfseries\cellcolor[gray]{0.8}70.82\% ~~1 & 77.57\% ~~3 & 75.85\% ~~4 & 8.28\% ~~3 & 85.05\% ~~3\\\hline
\hline
\it HIT-SCIR \tiny\cite{Che:Dou:Xu:19} & \it 90.41\% ~~2 & \it 70.85\% ~~3 & \it \bfseries69.86\% ~~1 & \it 77.61\% ~~2 & \it \bfseries\cellcolor[gray]{0.8}79.37\% ~~1 & \it \bfseries\cellcolor[gray]{0.8}12.40\% ~~1 & \it \bfseries\cellcolor[gray]{0.8}86.20\% ~~1\\\hline
\it SJTU--NICT \tiny\cite{Li:Zha:Zha:19} & \it \bfseries\cellcolor[gray]{0.8}91.50\% ~~1 & \it 71.24\% ~~2 & \it 68.73\% ~~2 & \it \bfseries\cellcolor[gray]{0.8}77.62\% ~~1 & \it 77.74\% ~~2 & \it 9.40\% ~~2 & \it 85.27\% ~~2\\\hline
\it SUDA--Alibaba \tiny\cite{Zha:Jia:Xia:19}\kern-.7em & \it 86.01\% ~~5 & \it 69.50\% ~~4 & \it 68.24\% ~~3 & \it 77.11\% ~~3 & \it 76.85\% ~~3 & \it 8.16\% ~~3 & \it 83.96\% ~~3\\\hline
\it Saarland \tiny\cite{Don:Fow:Gro:19} & \it 86.70\% ~~4 & \it \bfseries71.33\% ~~1 & \it 61.11\% ~~5 & \it 75.08\% ~~5 & \it 75.01\% ~~4 & \multicolumn{1}{c|}{---} & \it 81.87\% ~~4\\\hline

  \end{tabular}
  \caption{Overall results, macro-averaged on all frameworks. We present F1 scores and
  ranks compared to official ST submissions. Results with rank 1 are typeset
  in \textbf{bold}, best results in each column have \colorbox{gray!50}{gray background}.}
  \label{tbl:results}
\end{table*}

\begin{table*}[p]
  \centering
  \small\setlength{\tabcolsep}{4.4pt}
  
  System & \multicolumn{1}{c|}{Tops} & \multicolumn{1}{c|}{Labels} & \multicolumn{1}{c|}{Properties} & \multicolumn{1}{c|}{Anchors} & \multicolumn{1}{c|}{Edges} & \multicolumn{1}{c|}{Attributes} & \multicolumn{1}{c}{All}\\\hline\hline
Bugfix ST submission & 87.39\% ~~8 & \bfseries97.29\% ~~1 & 94.50\% ~~5 & 99.02\% ~~6 & 88.32\% ~~8 & \multicolumn{1}{c|}{---} & 94.66\% ~~5\\\hline
99\% training data & 88.36\% ~~8 & \bfseries97.38\% ~~1 & 94.57\% ~~5 & 99.04\% ~~6 & 88.47\% ~~8 & \multicolumn{1}{c|}{---} & 94.75\% ~~4\\\hline
No BERT embeddings & 80.70\% ~~9 & 96.24\% ~~2 & 92.19\% ~~7 & 98.45\% ~~8 & 80.06\% 10 & \multicolumn{1}{c|}{---} & 91.75\% ~~8\\\hline
Ensemble & 89.06\% ~~7 & \bfseries\cellcolor[gray]{0.8}97.51\% ~~1 & 94.86\% ~~4 & 99.12\% ~~3 & 89.72\% ~~8 & \multicolumn{1}{c|}{---} & 95.17\% ~~2\\\hline
\hline
\it HIT-SCIR \tiny\cite{Che:Dou:Xu:19} & \it 92.65\% ~~3 & \it 93.00\% ~~4 & \it 95.33\% ~~3 & \it \bfseries\cellcolor[gray]{0.8}99.28\% ~~1 & \it \cellcolor[gray]{0.8}92.54\% ~~2 & \multicolumn{1}{c|}{---} & \it 95.08\% ~~2\\\hline
\it SJTU--NICT \tiny\cite{Li:Zha:Zha:19} & \it \cellcolor[gray]{0.8}93.26\% ~~2 & \it 94.89\% ~~3 & \it \cellcolor[gray]{0.8}95.49\% ~~2 & \it 99.27\% ~~2 & \it 92.39\% ~~3 & \multicolumn{1}{c|}{---} & \it \bfseries\cellcolor[gray]{0.8}95.50\% ~~1\\\hline
\it SUDA--Alibaba \tiny\cite{Zha:Jia:Xia:19}\kern-.7em & \it 91.13\% ~~6 & \it 90.27\% ~~8 & \it 91.51\% ~~7 & \it 98.16\% ~~8 & \it 89.84\% ~~7 & \multicolumn{1}{c|}{---} & \it 92.26\% ~~7\\\hline
\it Saarland \tiny\cite{Don:Fow:Gro:19} & \it 85.87\% ~~8 & \it \bfseries96.82\% ~~1 & \it 93.55\% ~~5 & \it 99.05\% ~~5 & \it 90.95\% ~~6 & \multicolumn{1}{c|}{---} & \it 94.69\% ~~4\\\hline

  \noalign{\smallskip} \noalign{(a) DM framework}\noalign{\bigskip}
  System & \multicolumn{1}{c|}{Tops} & \multicolumn{1}{c|}{Labels} & \multicolumn{1}{c|}{Properties} & \multicolumn{1}{c|}{Anchors} & \multicolumn{1}{c|}{Edges} & \multicolumn{1}{c|}{Attributes} & \multicolumn{1}{c}{All}\\\hline\hline
Bugfix ST submission & 94.48\% ~~6 & \bfseries95.94\% ~~1 & 92.61\% ~~2 & 99.00\% ~~4 & 76.06\% ~~7 & \multicolumn{1}{c|}{---} & 90.96\% ~~4\\\hline
99\% training data & 86.49\% ~~9 & \bfseries96.05\% ~~1 & 92.70\% ~~2 & 99.00\% ~~3 & 76.37\% ~~7 & \multicolumn{1}{c|}{---} & 90.89\% ~~4\\\hline
No BERT embeddings & 67.57\% 12 & 95.14\% ~~3 & 90.72\% ~~7 & 98.47\% ~~8 & 68.22\% 10 & \multicolumn{1}{c|}{---} & 87.58\% ~~8\\\hline
Ensemble & 87.35\% ~~8 & \bfseries\cellcolor[gray]{0.8}96.19\% ~~1 & 93.04\% ~~2 & 99.02\% ~~3 & 78.20\% ~~7 & \multicolumn{1}{c|}{---} & \bfseries\cellcolor[gray]{0.8}91.51\% ~~1\\\hline
\hline
\it HIT-SCIR \tiny\cite{Che:Dou:Xu:19} & \it 96.03\% ~~3 & \it 89.30\% ~~5 & \it \bfseries\cellcolor[gray]{0.8}93.10\% ~~1 & \it \bfseries\cellcolor[gray]{0.8}99.12\% ~~1 & \it 79.65\% ~~3 & \multicolumn{1}{c|}{---} & \it 90.55\% ~~4\\\hline
\it SJTU--NICT \tiny\cite{Li:Zha:Zha:19} & \it \bfseries\cellcolor[gray]{0.8}96.30\% ~~1 & \it 93.14\% ~~4 & \it 91.57\% ~~5 & \it 99.11\% ~~2 & \it \bfseries\cellcolor[gray]{0.8}80.27\% ~~1 & \multicolumn{1}{c|}{---} & \it 91.19\% ~~3\\\hline
\it SUDA--Alibaba \tiny\cite{Zha:Jia:Xia:19}\kern-.7em & \it 86.55\% ~~8 & \it 84.51\% ~~8 & \it 85.03\% ~~8 & \it 97.51\% ~~8 & \it 75.22\% ~~7 & \multicolumn{1}{c|}{---} & \it 85.56\% ~~8\\\hline
\it Saarland \tiny\cite{Don:Fow:Gro:19} & \it 93.50\% ~~6 & \it 95.21\% ~~2 & \it 92.20\% ~~4 & \it 99.00\% ~~3 & \it 78.32\% ~~6 & \multicolumn{1}{c|}{---} & \it \bfseries91.28\% ~~1\\\hline

  \noalign{\smallskip} \noalign{(b) PSD framework}\noalign{\bigskip}
  System & \multicolumn{1}{c|}{Tops} & \multicolumn{1}{c|}{Labels} & \multicolumn{1}{c|}{Properties} & \multicolumn{1}{c|}{Anchors} & \multicolumn{1}{c|}{Edges} & \multicolumn{1}{c|}{Attributes} & \multicolumn{1}{c}{All}\\\hline\hline
Bugfix ST submission & 82.82\% ~~6 & 89.99\% ~~3 & \bfseries91.21\% ~~1 & 92.67\% ~~4 & 84.76\% ~~7 & \multicolumn{1}{c|}{---} & 89.12\% ~~4\\\hline
99\% training data & 83.79\% ~~6 & 90.19\% ~~3 & \bfseries91.19\% ~~1 & 92.88\% ~~4 & 85.09\% ~~6 & \multicolumn{1}{c|}{---} & 89.37\% ~~4\\\hline
No BERT embeddings & 73.91\% ~~8 & 84.52\% ~~5 & 85.76\% ~~3 & 89.08\% ~~5 & 76.73\% ~~7 & \multicolumn{1}{c|}{---} & 83.43\% ~~7\\\hline
Ensemble & 84.59\% ~~6 & 90.86\% ~~2 & \bfseries\cellcolor[gray]{0.8}92.00\% ~~1 & 93.52\% ~~3 & 86.55\% ~~6 & \multicolumn{1}{c|}{---} & 90.29\% ~~3\\\hline
\hline
\it HIT-SCIR \tiny\cite{Che:Dou:Xu:19} & \it 85.23\% ~~5 & \it 89.45\% ~~3 & \it 89.54\% ~~2 & \it 94.29\% ~~2 & \it 88.77\% ~~3 & \multicolumn{1}{c|}{---} & \it 90.75\% ~~2\\\hline
\it SJTU--NICT \tiny\cite{Li:Zha:Zha:19} & \it 87.72\% ~~3 & \it 89.42\% ~~4 & \it 77.53\% ~~4 & \it 93.37\% ~~3 & \it 87.82\% ~~4 & \multicolumn{1}{c|}{---} & \it 89.90\% ~~3\\\hline
\it SUDA--Alibaba \tiny\cite{Zha:Jia:Xia:19}\kern-.7em & \it \cellcolor[gray]{0.8}89.94\% ~~2 & \it \bfseries\cellcolor[gray]{0.8}91.20\% ~~1 & \it \bfseries89.72\% ~~1 & \it \bfseries\cellcolor[gray]{0.8}94.86\% ~~1 & \it 89.66\% ~~2 & \multicolumn{1}{c|}{---} & \it \bfseries\cellcolor[gray]{0.8}91.85\% ~~1\\\hline
\it Saarland \tiny\cite{Don:Fow:Gro:19} & \it 86.31\% ~~4 & \it 90.61\% ~~2 & \it 78.99\% ~~3 & \it 86.55\% ~~6 & \it \bfseries\cellcolor[gray]{0.8}90.96\% ~~1 & \multicolumn{1}{c|}{---} & \it 89.10\% ~~4\\\hline

  \noalign{\smallskip} \noalign{(c) EDS framework}\noalign{\bigskip}
  System & \multicolumn{1}{c|}{Tops} & \multicolumn{1}{c|}{Labels} & \multicolumn{1}{c|}{Properties} & \multicolumn{1}{c|}{Anchors} & \multicolumn{1}{c|}{Edges} & \multicolumn{1}{c|}{Attributes} & \multicolumn{1}{c}{All}\\\hline\hline
Bugfix ST submission & 62.51\% ~~9 & \multicolumn{1}{c|}{---} & \multicolumn{1}{c|}{---} & 95.44\% ~~2 & 59.45\% ~~4 & 39.35\% ~~4 & 73.24\% ~~4\\\hline
99\% training data & 63.53\% ~~9 & \multicolumn{1}{c|}{---} & \multicolumn{1}{c|}{---} & 95.80\% ~~2 & 60.51\% ~~4 & 39.81\% ~~4 & 73.95\% ~~4\\\hline
No BERT embeddings & 59.40\% 10 & \multicolumn{1}{c|}{---} & \multicolumn{1}{c|}{---} & 94.11\% ~~5 & 48.70\% ~~8 & 26.52\% ~~6 & 66.90\% ~~7\\\hline
Ensemble & 63.28\% ~~9 & \multicolumn{1}{c|}{---} & \multicolumn{1}{c|}{---} & 96.19\% ~~2 & 62.14\% ~~4 & 41.39\% ~~3 & 75.22\% ~~4\\\hline
\hline
\it HIT-SCIR \tiny\cite{Che:Dou:Xu:19} & \it \bfseries\cellcolor[gray]{0.8}\llap{1}00.00\% ~~1 & \multicolumn{1}{c|}{---} & \multicolumn{1}{c|}{---} & \it 95.36\% ~~3 & \it \bfseries\cellcolor[gray]{0.8}72.66\% ~~1 & \it \bfseries\cellcolor[gray]{0.8}61.98\% ~~1 & \it \bfseries\cellcolor[gray]{0.8}81.67\% ~~1\\\hline
\it SJTU--NICT \tiny\cite{Li:Zha:Zha:19} & \it 95.31\% ~~5 & \multicolumn{1}{c|}{---} & \multicolumn{1}{c|}{---} & \it \bfseries\cellcolor[gray]{0.8}96.36\% ~~1 & \it 65.56\% ~~3 & \it 47.00\% ~~2 & \it 77.80\% ~~3\\\hline
\it SUDA--Alibaba \tiny\cite{Zha:Jia:Xia:19}\kern-.7em & \it 99.56\% ~~3 & \multicolumn{1}{c|}{---} & \multicolumn{1}{c|}{---} & \it 95.02\% ~~4 & \it 67.74\% ~~2 & \it 40.80\% ~~3 & \it 78.43\% ~~2\\\hline
\it Saarland \tiny\cite{Don:Fow:Gro:19} & \it 80.95\% ~~8 & \multicolumn{1}{c|}{---} & \multicolumn{1}{c|}{---} & \it 90.81\% ~~6 & \it 52.66\% ~~6 & \multicolumn{1}{c|}{---} & \it 67.55\% ~~6\\\hline

  \noalign{\smallskip} \noalign{(d) UCCA framework}\noalign{\bigskip}
  System & \multicolumn{1}{c|}{Tops} & \multicolumn{1}{c|}{Labels} & \multicolumn{1}{c|}{Properties} & \multicolumn{1}{c|}{Anchors} & \multicolumn{1}{c|}{Edges} & \multicolumn{1}{c|}{Attributes} & \multicolumn{1}{c}{All}\\\hline\hline
Bugfix ST submission & 80.17\% ~~6 & 82.09\% ~~4 & 71.44\% ~~5 & \multicolumn{1}{c|}{---} & 60.83\% ~~6 & \multicolumn{1}{c|}{---} & 71.83\% ~~4\\\hline
99\% training data & 80.77\% ~~6 & 81.69\% ~~4 & 72.45\% ~~4 & \multicolumn{1}{c|}{---} & 60.93\% ~~6 & \multicolumn{1}{c|}{---} & 71.73\% ~~5\\\hline
No BERT embeddings & 70.91\% ~~8 & 77.67\% ~~6 & 66.36\% ~~6 & \multicolumn{1}{c|}{---} & 51.39\% ~~8 & \multicolumn{1}{c|}{---} & 65.29\% ~~7\\\hline
Ensemble & 81.39\% ~~6 & 82.40\% ~~3 & 74.21\% ~~4 & \multicolumn{1}{c|}{---} & 62.65\% ~~3 & \multicolumn{1}{c|}{---} & \cellcolor[gray]{0.8}73.03\% ~~2\\\hline
\hline
\it HIT-SCIR \tiny\cite{Che:Dou:Xu:19} & \it 78.15\% ~~7 & \it \cellcolor[gray]{0.8}82.51\% ~~2 & \it 71.33\% ~~5 & \multicolumn{1}{c|}{---} & \it \cellcolor[gray]{0.8}63.21\% ~~2 & \multicolumn{1}{c|}{---} & \it 72.94\% ~~2\\\hline
\it SJTU--NICT \tiny\cite{Li:Zha:Zha:19} & \it 84.88\% ~~4 & \it 78.78\% ~~5 & \it \bfseries\cellcolor[gray]{0.8}79.08\% ~~1 & \multicolumn{1}{c|}{---} & \it 62.64\% ~~3 & \multicolumn{1}{c|}{---} & \it 71.97\% ~~3\\\hline
\it SUDA--Alibaba \tiny\cite{Zha:Jia:Xia:19}\kern-.7em & \it 62.86\% ~~9 & \it 81.53\% ~~4 & \it 74.96\% ~~3 & \multicolumn{1}{c|}{---} & \it 61.78\% ~~5 & \multicolumn{1}{c|}{---} & \it 71.72\% ~~5\\\hline
\it Saarland \tiny\cite{Don:Fow:Gro:19} & \it \bfseries\cellcolor[gray]{0.8}86.89\% ~~1 & \it 74.02\% ~~6 & \it 40.79\% ~~7 & \multicolumn{1}{c|}{---} & \it 62.16\% ~~4 & \multicolumn{1}{c|}{---} & \it 66.72\% ~~6\\\hline

  \noalign{\smallskip} \noalign{(e) AMR framework}\noalign{\bigskip}
  \end{tabular}
  \caption{Results on individual frameworks. We present F1 scores and
  ranks compared to official ST submissions. Results with rank 1 are typeset
  in \textbf{bold}, best results in each column have \colorbox{gray!50}{gray background}.}
  \label{tbl:all_results}
\end{table*}

We present the overall results of our system in Table~\ref{tbl:results}. Please
note that our official shared task submission contained an error -- test data
companion analyses had been updated during the evaluation phase, but we used
the original incorrect ones for DM, PSD and EDS frameworks. The error was
discovered only after the official deadline, at which point we sent a bugfix
submission using the same trained models, the only difference being the
utilization of the correct test data analyses during prediction. We present
both these submissions in the Table~\ref{tbl:results}, but refer only to the
bugfix submission from now on.

The overall results of our system using the official MRP metric are present
in Table~\ref{tbl:results}. All reported scores are macro-averaged F1 scores
of all five frameworks. The results for individual frameworks are presented
in Table~\ref{tbl:all_results}.

Our bugfix submission would score third in in the macro-averaged \textit{all}
metric. Overall, our system reaches high accuracy in node \textit{labels} and
\textit{properties} prediction, ranking first in both of them. These
predictions employ the relative encoding extended from UDPipe and demonstrate
its effectiveness.

The weakest points of our system are the \textit{top} nodes prediction and
\textit{edges} prediction. We hypothesise that the lower performance of the
\textbf{AddEdges} operation could be improved by better node representation
(i.e., including all dependent tokens of a node, not only the one token generating the
node) and by a better edge prediction architecture (i.e., global decision over edge connection
in the context of all graph nodes instead of considering only the current node
pair).

Framework-wise, our system would achieve ranks 5, 4, 4, 4 and 4 on DM, PSD, EDS,
UCCA and AMR, respectively, showing relatively balanced performance. The
largest absolute performance gap of our system occurs on UCCA, where we reach
8 percent points lower score than the best system, which is supposedly
caused by the fact that there are no \textit{labels} and \textit{properties}
which our system excels in predicting, and also by the constituency structure
of the UCCA graphs which we represent poorly.

\subsection{Ablation Experiments}
\label{sec:ablation}

Given that our submission utilized only 90\% of the available
training data, we also evaluated a variant employing 99\% of the training data,
keeping the last 1\% for error detection. However, as
Tables~\ref{tbl:results}~and~\ref{tbl:all_results} show, the results are nearly
identical.

In order to asses the BERT embeddings effect, we further evaluated
a version of our system without them. The macro-averaged \textit{all} performance
without BERT embeddings is substantially lower, 79\% compared to 84\%. Generally
all metrics decrease without BERT embeddings, showing that contextual
embeddings help ``everywhere''.

Lastly, we evaluated performance of an 5-model ensemble. Each model was trained
using 99\% of the training data and utilized different random initialization.
The system performance increased by more than 1 percent point. Although the
overall rank of the ensemble is unchanged, the rank on individual frameworks
increased from 5 to 2 on DM, from 4 to 1 on PSD, 4 to 3 on EDS and 4 to 2 on
AMR. As with the non-ensemble system, the weakest point of our solution
are the \textit{edge} predictions, which rank 8, 7, 6, 4 and 3 on DM, PSD, EDS,
UCCA and AMR, respectively.

\section{Conclusions}

We introduced a uniform graph-to-graph architecture for parsing
into semantic graphs. The model implicitly learns the linguistic information
and the graph structure without the need for any specific hand-crafted or
structural knowledge and is suitable for any directed graph, including graphs
with cycles. In contrast to a transition-based system, we build the graph in
a layer-wise fashion, with operations joined in groups.

\section*{Acknowledgments}

The work described herein has been supported by OP VVV VI LINDAT/CLARIN project
(CZ.02.1.01/0.0/0.0/16\_013/0001781) and it has been supported and has been
using language resources developed by the LINDAT/CLARIN project (LM2015071) of
the Ministry of Education, Youth and Sports of the Czech Republic.

We thank the anonymous reviewers for their insightful comments.

\bibliography{mrp2019,mrp,sdp}

\providecommand{\fromto}[2]{#1$\,$--$\,$#2}\providecommand{\pages}[3]{#2$\,$--$\,$#3}\providecommand\Beek[1]{\mbox{#1}}\providecommand\Noord[1]{\mbox{#1}}\providecommand\Lohuizen[1]{\mbox{#1}}\providecommand{\fromto}[2]{#1$\,$--$\,$#2}
\begin{thebibliography}{28}
\expandafter\ifx\csname natexlab\endcsname\relax\def\natexlab#1{#1}\fi

\bibitem[{Abend and Rappoport(2013)}]{UCCA2013}
Omri Abend and Ari Rappoport. 2013.
\newblock \href {https://www.aclweb.org/anthology/P13-1023} {Universal
  conceptual cognitive annotation ({UCCA})}.
\newblock In \emph{Proceedings of the 51st Annual Meeting of the Association
  for Computational Linguistics (Volume 1: Long Papers)}, pages 228--238,
  Sofia, Bulgaria. Association for Computational Linguistics.

\bibitem[{{Agrawal} et~al.(2019){Agrawal}, {Naresh Modi}, {Passos}, {Lavoie},
  {Agarwal}, {Shankar}, {Ganichev}, {Levenberg}, {Hong}, {Monga}, and
  {Cai}}]{tensorflow_eager}
Akshay {Agrawal}, Akshay {Naresh Modi}, Alexandre {Passos}, Allen {Lavoie},
  Ashish {Agarwal}, Asim {Shankar}, Igor {Ganichev}, Josh {Levenberg},
  Mingsheng {Hong}, Rajat {Monga}, and Shanqing {Cai}. 2019.
\newblock \href {http://arxiv.org/abs/1903.01855} {{TensorFlow Eager: A
  Multi-Stage, Python-Embedded DSL for Machine Learning}}.
\newblock \emph{arXiv e-prints}, page arXiv:1903.01855.

\bibitem[{Bahdanau et~al.(2014)Bahdanau, Cho, and Bengio}]{bahdanau14}
Dzmitry Bahdanau, Kyunghyun Cho, and Yoshua Bengio. 2014.
\newblock \href {http://arxiv.org/abs/1409.0473} {Neural machine translation by
  jointly learning to align and translate}.
\newblock \emph{CoRR}, abs/1409.0473.

\bibitem[{Banarescu et~al.(2013)Banarescu, Bonial, Cai, Georgescu, Griffitt,
  Hermjakob, Knight, Koehn, Palmer, and Schneider}]{AMR2013}
Laura Banarescu, Claire Bonial, Shu Cai, Madalina Georgescu, Kira Griffitt, Ulf
  Hermjakob, Kevin Knight, Philipp Koehn, Martha Palmer, and Nathan Schneider.
  2013.
\newblock \href {https://www.aclweb.org/anthology/W13-2322} {Abstract meaning
  representation for sembanking}.
\newblock In \emph{Proceedings of the 7th Linguistic Annotation Workshop and
  Interoperability with Discourse}, pages 178--186, Sofia, Bulgaria.
  Association for Computational Linguistics.

\bibitem[{Che et~al.(2019)Che, Dou, Xu, Wang, Liu, and Liu}]{Che:Dou:Xu:19}
Wanxiang Che, Longxu Dou, Yang Xu, Yuxuan Wang, Yijia Liu, and Ting Liu. 2019.
\newblock {HIT-SCIR} at {MRP}~2019: {A} unified pipeline for meaning
  representation parsing via efficient training and effective encoding.
\newblock In \emph{Proceedings of the Shared Task on Cross-Framework Meaning
  Representation Parsing at the 2019 {C}onference on {N}atural {L}anguage
  {L}earning}, pages \pages{--}{76}{85}, Hong Kong, China.

\bibitem[{Devlin et~al.(2019)Devlin, Chang, Lee, and Toutanova}]{BERT2019}
Jacob Devlin, Ming-Wei Chang, Kenton Lee, and Kristina Toutanova. 2019.
\newblock \href {https://doi.org/10.18653/v1/N19-1423} {{BERT}: Pre-training of
  deep bidirectional transformers for language understanding}.
\newblock In \emph{Proceedings of the 2019 Conference of the North {A}merican
  Chapter of the Association for Computational Linguistics: Human Language
  Technologies, Volume 1 (Long and Short Papers)}, pages 4171--4186,
  Minneapolis, Minnesota. Association for Computational Linguistics.

\bibitem[{Donatelli et~al.(2019)Donatelli, Fowlie, Groschwitz, Koller,
  Lindemann, Mina, and Weißenhorn}]{Don:Fow:Gro:19}
Lucia Donatelli, Meaghan Fowlie, Jonas Groschwitz, Alexander Koller, Matthias
  Lindemann, Mario Mina, and Pia Weißenhorn. 2019.
\newblock Saarland at {MRP}~2019: {C}ompositional parsing across all
  graphbanks.
\newblock In \emph{Proceedings of the Shared Task on Cross-Framework Meaning
  Representation Parsing at the 2019 {C}onference on {N}atural {L}anguage
  {L}earning}, pages \pages{--}{66}{75}, Hong Kong, China.

\bibitem[{Flanigan et~al.(2016)Flanigan, Dyer, Smith, and Carbonell}]{JAMR2016}
Jeffrey Flanigan, Chris Dyer, Noah~A. Smith, and Jaime Carbonell. 2016.
\newblock \href {https://doi.org/10.18653/v1/S16-1186} {{CMU} at
  {S}em{E}val-2016 task 8: Graph-based {AMR} parsing with infinite ramp loss}.
\newblock In \emph{Proceedings of the 10th International Workshop on Semantic
  Evaluation (SemEval-2016)}, pages 1202--1206, San Diego, California.
  Association for Computational Linguistics.

\bibitem[{Haji{\v{c}} et~al.(2012)Haji{\v{c}}, Haji{\v{c}}ov{\'a},
  Panevov{\'a}, Sgall, Bojar, Cinkov{\'a}, Fu{\v{c}}{\'\i}kov{\'a},
  Mikulov{\'a}, Pajas, Popelka, Semeck{\'y}, {\v{S}}indlerov{\'a},
  {\v{S}}t{\v{e}}p{\'a}nek, Toman, Ure{\v{s}}ov{\'a}, and
  {\v{Z}}abokrtsk{\'y}}]{Hajic2012}
Jan Haji{\v{c}}, Eva Haji{\v{c}}ov{\'a}, Jarmila Panevov{\'a}, Petr Sgall,
  Ond{\v{r}}ej Bojar, Silvie Cinkov{\'a}, Eva Fu{\v{c}}{\'\i}kov{\'a}, Marie
  Mikulov{\'a}, Petr Pajas, Jan Popelka, Ji{\v{r}}{\'\i} Semeck{\'y}, Jana
  {\v{S}}indlerov{\'a}, Jan {\v{S}}t{\v{e}}p{\'a}nek, Josef Toman, Zde{\v{n}}ka
  Ure{\v{s}}ov{\'a}, and Zden{\v{e}}k {\v{Z}}abokrtsk{\'y}. 2012.
\newblock \href
  {http://www.lrec-conf.org/proceedings/lrec2012/pdf/510_Paper.pdf} {Announcing
  {P}rague {C}zech-{E}nglish dependency treebank 2.0}.
\newblock In \emph{Proceedings of the Eighth International Conference on
  Language Resources and Evaluation ({LREC}-2012)}, pages 3153--3160, Istanbul,
  Turkey. European Languages Resources Association (ELRA).

\bibitem[{Hershcovich et~al.(2018)Hershcovich, Abend, and
  Rappoport}]{Hershovich2018}
Daniel Hershcovich, Omri Abend, and Ari Rappoport. 2018.
\newblock \href {https://doi.org/10.18653/v1/P18-1035} {Multitask parsing
  across semantic representations}.
\newblock In \emph{Proceedings of the 56th Annual Meeting of the Association
  for Computational Linguistics (Volume 1: Long Papers)}, pages 373--385,
  Melbourne, Australia. Association for Computational Linguistics.

\bibitem[{Hershcovich et~al.(2019)Hershcovich, Aizenbud, Choshen, Sulem,
  Rappoport, and Abend}]{SemEval2019}
Daniel Hershcovich, Zohar Aizenbud, Leshem Choshen, Elior Sulem, Ari Rappoport,
  and Omri Abend. 2019.
\newblock \href {https://doi.org/10.18653/v1/S19-2001} {{S}em{E}val-2019 task
  1: Cross-lingual semantic parsing with {UCCA}}.
\newblock In \emph{Proceedings of the 13th International Workshop on Semantic
  Evaluation}, pages 1--10, Minneapolis, Minnesota, USA. Association for
  Computational Linguistics.

\bibitem[{Ivanova et~al.(2012)Ivanova, Oepen, {{\O}}vrelid, and
  Flickinger}]{DM2012}
Angelina Ivanova, Stephan Oepen, Lilja {{\O}}vrelid, and Dan Flickinger. 2012.
\newblock \href {http://dl.acm.org/citation.cfm?id=2392747.2392751} {Who did
  what to whom?: A contrastive study of syntacto-semantic dependencies}.
\newblock In \emph{Proceedings of the Sixth Linguistic Annotation Workshop},
  LAW VI '12, pages 2--11, Stroudsburg, PA, USA. Association for Computational
  Linguistics.

\bibitem[{Kingma and Ba(2014)}]{Adam}
Diederik Kingma and Jimmy Ba. 2014.
\newblock {Adam: A Method for Stochastic Optimization}.
\newblock \emph{International Conference on Learning Representations}.

\bibitem[{Li et~al.(2019)Li, Zhao, Zhang, Wang, Utiyama, and
  Sumita}]{Li:Zha:Zha:19}
Zuchao Li, Hai Zhao, Zhuosheng Zhang, Rui Wang, Masao Utiyama, and Eiichiro
  Sumita. 2019.
\newblock {SJTU--NICT} at {MRP}~2019: {M}ulti-task learning for end-to-end
  uniform semantic graph parsing.
\newblock In \emph{Proceedings of the Shared Task on Cross-Framework Meaning
  Representation Parsing at the 2019 {C}onference on {N}atural {L}anguage
  {L}earning}, pages \pages{--}{45}{54}, Hong Kong, China.

\bibitem[{Ling et~al.(2015)Ling, Lu{\'{i}}s, Marujo, Astudillo, Amir, Dyer,
  Black, and Trancoso}]{Ling2015}
Wang Ling, Tiago Lu{\'{i}}s, Lu{\'{i}}s Marujo, Ram{\'{o}}n~Fernandez
  Astudillo, Silvio Amir, Chris Dyer, Alan~W. Black, and Isabel Trancoso. 2015.
\newblock Finding function in form: Compositional character models for open
  vocabulary word representation.
\newblock \emph{CoRR}.

\bibitem[{May(2016)}]{SemEval2016}
Jonathan May. 2016.
\newblock \href {https://www.aclweb.org/anthology/S16-1166/} {{S}em{E}val-2016
  task 8: Meaning representation parsing}.
\newblock In \emph{Proceedings of the 10th International Workshop on Semantic
  Evaluation, SemEval@NAACL-HLT 2016, San Diego, CA, USA, June 16-17, 2016},
  pages 1063--1073. The Association for Computer Linguistics.

\bibitem[{May and Priyadarshi(2017)}]{SemEval2017}
Jonathan May and Jay Priyadarshi. 2017.
\newblock \href {https://doi.org/10.18653/v1/S17-2090} {{S}em{E}val-2017 task
  9: Abstract meaning representation parsing and generation}.
\newblock In \emph{Proceedings of the 11th International Workshop on Semantic
  Evaluation ({S}em{E}val-2017)}, pages 536--545, Vancouver, Canada.
  Association for Computational Linguistics.

\bibitem[{Mikolov et~al.(2018)Mikolov, Grave, Bojanowski, Puhrsch, and
  Joulin}]{FastText2018}
Tomas Mikolov, Edouard Grave, Piotr Bojanowski, Christian Puhrsch, and Armand
  Joulin. 2018.
\newblock {Advances in Pre-Training Distributed Word Representations}.
\newblock In \emph{Proceedings of the International Conference on Language
  Resources and Evaluation (LREC 2018)}.

\bibitem[{Miyao et~al.(2014)Miyao, Oepen, and Zeman}]{Miy:Oep:Zem:14}
Yusuke Miyao, Stephan Oepen, and Daniel Zeman. 2014.
\newblock In-{H}ouse. {A}n ensemble of pre-existing off-the-shelf parsers.
\newblock In \emph{Proceedings of the 8th {I}nternational {W}orkshop on
  {S}emantic {E}valuation}, page \fromto{63}{72}, Dublin, Ireland.

\bibitem[{Oepen et~al.(2019)Oepen, Abend, Haji\v{c}, Hershcovich, Kuhlmann,
  O'Gorman, Xue, Chun, Straka, and Ure\v{s}ov{\'a}}]{Oep:Abe:Haj:19}
Stephan Oepen, Omri Abend, Jan Haji\v{c}, Daniel Hershcovich, Marco Kuhlmann,
  Tim O'Gorman, Nianwen Xue, Jayeol Chun, Milan Straka, and Zde\v{n}ka
  Ure\v{s}ov{\'a}. 2019.
\newblock {MRP}~2019: {C}ross-framework {M}eaning {R}epresentation {P}arsing.
\newblock In \emph{Proceedings of the Shared Task on Cross-Framework Meaning
  Representation Parsing at the 2019 {C}onference on {N}atural {L}anguage
  {L}earning}, pages \pages{--}{1}{27}, Hong Kong, China.

\bibitem[{Oepen and L{\o}nning(2006)}]{Oepen2006}
Stephan Oepen and Jan~Tore L{\o}nning. 2006.
\newblock \href {http://www.lrec-conf.org/proceedings/lrec2006/pdf/364_pdf.pdf}
  {Discriminant-based {MRS} banking}.
\newblock In \emph{Proceedings of the Fifth International Conference on
  Language Resources and Evaluation ({LREC}{'}06)}, Genoa, Italy. European
  Language Resources Association (ELRA).

\bibitem[{Ratinov and Roth(2009)}]{RatinovRo09}
Lev Ratinov and Dan Roth. 2009.
\newblock \href {http://cogcomp.org/papers/RatinovRo09.pdf} {{Design Challenges
  and Misconceptions in Named Entity Recognition}}.
\newblock In \emph{Proc. of the Conference on Computational Natural Language
  Learning (CoNLL)}.

\bibitem[{Straka(2018)}]{UDPipe2018}
Milan Straka. 2018.
\newblock {UDPipe 2.0 Prototype at CoNLL 2018 UD Shared Task}.
\newblock In \emph{Proceedings of CoNLL 2018: The SIGNLL Conference on
  Computational Natural Language Learning}, pages 197--207, Stroudsburg, PA,
  USA. Association for Computational Linguistics.

\bibitem[{Straka et~al.(2019)Straka, Strakov{\'a}, and
  Hajic}]{UDPipe-SIGMORPHON2019}
Milan Straka, Jana Strakov{\'a}, and Jan Hajic. 2019.
\newblock \href {https://www.aclweb.org/anthology/W19-4212} {{UDPipe at
  SIGMORPHON 2019: Contextualized Embeddings, Regularization with Morphological
  Categories, Corpora Merging}}.
\newblock In \emph{Proceedings of the 16th Workshop on Computational Research
  in Phonetics, Phonology, and Morphology}, pages 95--103, Florence, Italy.
  Association for Computational Linguistics.

\bibitem[{{Straka} et~al.(2019){Straka}, {Strakov{\'a}}, and
  {Haji{\v{c}}}}]{UDPipeBERT2019}
Milan {Straka}, Jana {Strakov{\'a}}, and Jan {Haji{\v{c}}}. 2019.
\newblock \href {http://arxiv.org/abs/1908.07448} {{Evaluating Contextualized
  Embeddings on 54 Languages in POS Tagging, Lemmatization and Dependency
  Parsing}}.
\newblock \emph{arXiv e-prints}, page arXiv:1908.07448.

\bibitem[{Zeman et~al.(2018)Zeman, Haji{\v{c}}, Popel, Potthast, Straka,
  Ginter, Nivre, and Petrov}]{udst2018:overview}
Daniel Zeman, Jan Haji{\v{c}}, Martin Popel, Martin Potthast, Milan Straka,
  Filip Ginter, Joakim Nivre, and Slav Petrov. 2018.
\newblock {CoNLL 2018 Shared Task: Multilingual Parsing from Raw Text to
  Universal Dependencies}.
\newblock In \emph{Proceedings of the CoNLL 2018 Shared Task: Multilingual
  Parsing from Raw Text to Universal Dependencies}, pages 1--20, Brussels,
  Belgium. Association for Computational Linguistics.

\bibitem[{Zhang et~al.(2019{\natexlab{a}})Zhang, Ma, Duh, and
  Van~Durme}]{Zhang2019}
Sheng Zhang, Xutai Ma, Kevin Duh, and Benjamin Van~Durme. 2019{\natexlab{a}}.
\newblock \href {https://doi.org/10.18653/v1/P19-1009} {{AMR} parsing as
  sequence-to-graph transduction}.
\newblock In \emph{Proceedings of the 57th Annual Meeting of the Association
  for Computational Linguistics}, pages 80--94, Florence, Italy. Association
  for Computational Linguistics.

\bibitem[{Zhang et~al.(2019{\natexlab{b}})Zhang, Jiang, Xia, Cao, Wang, Li, and
  Zhang}]{Zha:Jia:Xia:19}
Yue Zhang, Wei Jiang, Qingrong Xia, Junjie Cao, Rui Wang, Zhenghua Li, and Min
  Zhang. 2019{\natexlab{b}}.
\newblock Suda--alibaba at {MRP}~2019: {G}raph-based models with {BERT}.
\newblock In \emph{Proceedings of the Shared Task on Cross-Framework Meaning
  Representation Parsing at the 2019 {C}onference on {N}atural {L}anguage
  {L}earning}, pages \pages{--}{149}{157}, Hong Kong, China.

\end{thebibliography}
\bibliographystyle{acl_natbib}

\end{document}